# Femural Head Autosegmentation for 3D Radiotherapy Planning: Preliminary Results

Bruno A. G. da Silva, Alvaro L. Fazenda and Fabiano C. Paixão*

*Abstract*—Contouring of organs at risk is an important but time consuming part of radiotherapy treatment planning. Several authors proposed methods for automatic delineation but the clinical expert's eye remains the gold standard method. In this paper, we present a totally visual software for automated delineation of the femural head. The software was successfully characterized in pelvic CT Scan of prostate patients (n=11). The automatic delineation was compared with manual and approved delineation through blind test evaluated by a panel of seniors radiation oncologists (n=9). Clinical experts evaluated that no any contouring correction were need in 77.8% and 67.8% of manual and automatic delineation respectively. Our results show that the software is robust, the automated delineation was reproducible in all patient, and its performance was similar to manually delineation.

## I. Introduction

DELINEATION of target volumes and multiple organs at risk are hugely important in external beam radiation therapy due the current advances achieved in the treatment delivery to the oncology patients [1,2]. The innovations of three-dimensional (3D) conformal radiotherapy and intensity-modulated radiotherapy (IMRT) have allowed radiation therapy dose distributions to be shaped much more closely to targets, allowing better avoidance of organs at risk and hence reduced normal tissue toxicity, and facilitating dose escalation to the target volume [1].

Organ delineation is currently the most time-consuming part in radiotherapy planning [3]. This procedure is usually performed by manual contouring in 2-D slices of CT images using drawing tools, and it may take several hours to delineate all structures of interest in a 3-D dataset by oncologists [3,4].

Manuscript was submitted to arXiv on December 07, 2018. This work was supported in part by the Federal University of Sao Paulo, AC Camargo Cancer Center, Sao Paulo Hospital, Technology Park of Sao Jose dos Campos, Varian Medical Systems, and Brazilian Ministry of Health. *Asterisk indicates corresponding author*

B.A.G. da Silva is with the Science and Technology Department, Federal University of Sao Paulo, Sao Jose dos Campos, SP 12.231-280 BRA, (e-mail: braggi.gomes@gmail.com).
Alvaro L. Fazenda is with the Science and Technology Department, Federal University of Sao Paulo, Sao Jose dos Campos, SP 12.231-280 BRA, (e-mail: alvaro.fazenda@unifesp.br).
F.C. Paixão, was with Pontifical Catholic University of Rio Grande do Sul, Porto Alegre, RS 90619-900 BRA. He is now with the Science and Technology Department, Federal University of Sao Paulo, Sao Jose dos Campos, SP 12.231-280 BRA, (e-mail: fcpaixao@unifesp.br)*.

However, manual contouring is considered the gold standards in radiotherapy planning, delineation of regions of interest (ROIs) still result in intra- and inter-observer variations due the factors such as user preferences and experience of the radiation oncologists [5,6].

Accordingly, it is clinically desirable to develop a robust, accurate and automatic algorithm for the segmentation in order to avoid the intra- and inter-observer variations and to reduce time-consuming and labor-intensive process of manual delineation [6,7]. Many methods have been proposed to address the aforementioned challenges such as deformable models, deformable registration, multi-atlas-based labeling, and deep learning [7,8] for all body region such as head and neck, lung, breast, abdomen, and pelvic.

The aim of this work were to develop a feasible, robust, fast, and automated algorithm for auto-segmentation of organs at risks for prostate cancer radiotherapy planning. The headspring algorithm developed was applied to delineate the femural head in pelvic region and the results were compared with manual and approved delineation through blind test evaluated by radiation oncologists.

## II. Method

*A. Auto-segmentation Software*
  *1. Overview*

An Application Program Interface (API) as a set of routines, protocols, and tools was developed in C# programming language for auto-segmentation software as a totally visual platform without the need to compile numerous individual algorithms. The software embedded consisted of 13 buttons classes to evaluate the most effectives image parameters during the research segmentation procedure. The visual buttons carry out functions such as browse file, show image, filter image, remove couch, point operation, wavelets, edge detector, sharpness, morphologic, blob detection, segmentation, save images, and overlap contour.

Segmented by auxiliary, preprocessing, and postprocessing, the parameters to achieve the end segmentation attained by the software could be totally customizable by the researcher and the final algorithm was the composition of steps from the best results generated by the appropriated parameters chosen. Within the software, functionalities capable of upload DICOM format images as input data and delineated ROIs images as output data has been created, as well as, image treatment from basic point of view, such as point operations, to complex algorithm, as Watershed [9] has been implemented. The figure 1 shows the screenshot of the software developed.

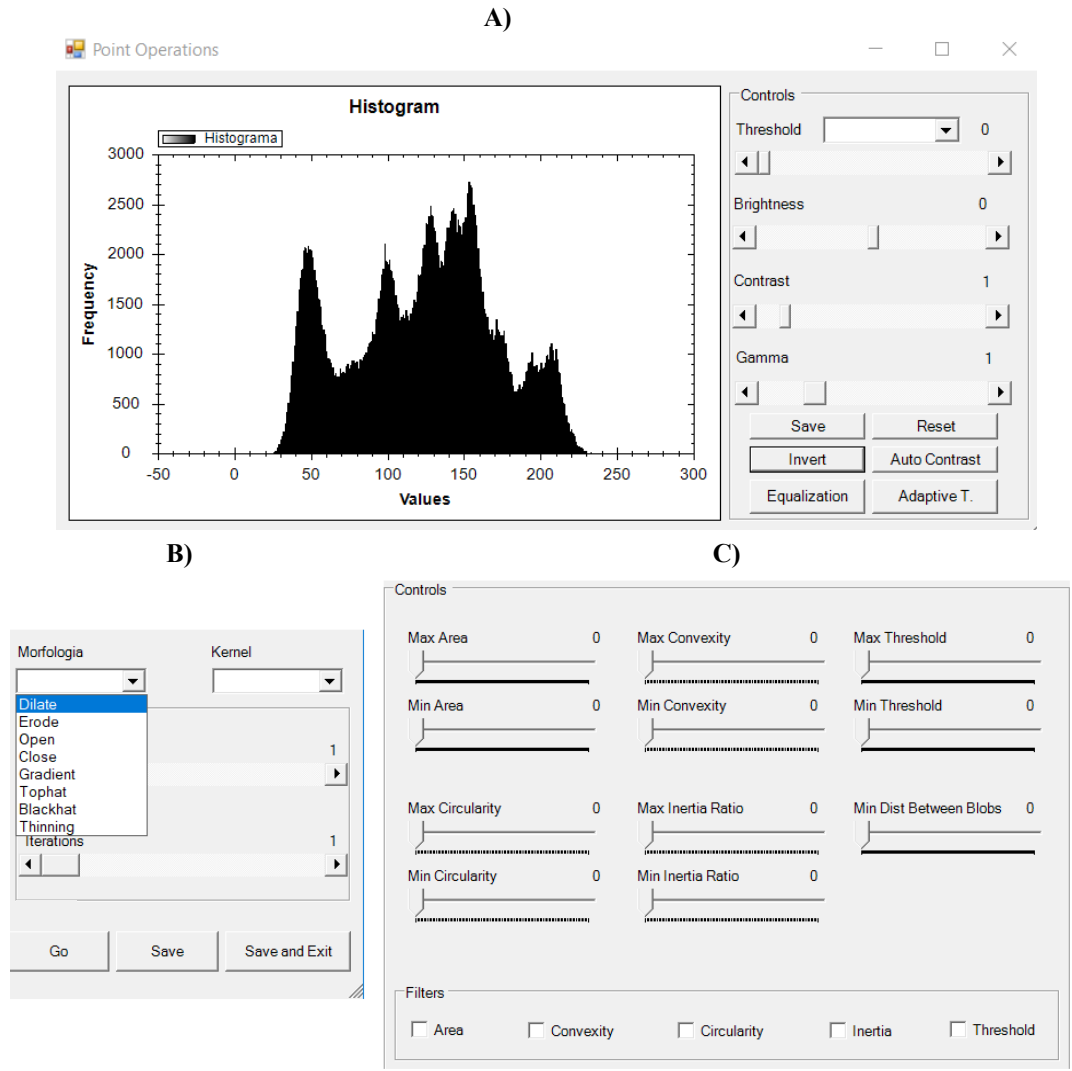

Fig. 1. Screenshot of the software developed for automatic segmentation of the femoral head in the 3D Radiotherapy Treatment Planning: A) Point Operations Parameters; B) Morphological Operations and its options; and C) Blob Detector and some types of filters as options.

*2. Pre-processing*

Different tools such as filters, point operations, wavelets, edge detector, and sharpness were implemented in this stage to draw up the image to be segmented. The preprocessing tools inserted within the software are described below.

*Filters:* Anisotropic Filtering [10], K-means [11], and Mean Shift [12] were appropriately used and selected to separate image noise from ROIs of each image of pelvic CT scan.

*Point Operations:* The intensity image was adjusted by operations such as: Simple, Local, Adaptative, and Otsu's thresholding methods; Brightness; Contrast; Gamma; Invert intensities; and Histogram equalization (figure 1a).

*Wavelets:* Additional image denoising was implemented using wavelet thresholding by focused on statistical modelling of wavelet coefficients and the optimal choice of thresholds [13].

*Edge Detector:* The border image, characterized by rapid acute change in image intensity, was detect runs different methods like Canny, Sobel, Prewitt, Laplace, and Hough Transform [14] with customized parameters.

*Sharpness:* It enhances image edges before any detection based on intensity variance by a Gaussian blur or Laplace operation [15].

*3. Post-processing*

The post-processing software structure was composed by several segmentation methods for extracting ROIs from background images. The postprocessing tools inserted within the software are described below.

*Morphological Operations:* The method pursues the goals of removing the imperfections on binary images by accounting for the form and structure of the image [16]. Dilate, Erode, Open, Close, Gradient, Tophat, Blackhat, and Thinning were implemented (fig. 1b).

*Blob Detector:* The blob is defined as a compact region lighter (or darker) than its background surrounded by a smoothly curved edged. It can be detected using cooperating relaxation processes to enhance their interior and edge probabilities [17]. The method implemented consisted in detecting group of connected pixels that share common property, such as color, size or shape via circularity, convexity or inertia ratio by the following steps: Thresholding, Grouping, Merging, and Center calculation.

*Floodfill:* The method determines the area connected to a given node in a multi-dimensional array [18]. It was used to fills with white pixels a closed edge area given information such as coordinate.

*Find Contours:* The method was used to find every closed contour from an ordered list classified by its area and returns the isolated contour as new image [19].

*Connected Components:* It is used in computer vision to detect connected regions in binary digital images, color images, and data with higher-dimensionality [20]. The method was used given an area for filters and shows all connected contours coordinates.

*Watershed:* The method is a powerful mathematical morphological tool for image segmentation [9] and it is popular in biomedical and medical image processing. Watershed method was implemented and parameters such as number of seeds and coordinates could be customized by user during the segmentation procedure.

*ORB Detector:* ORB is a fast robust local feature detector that can be applied in computer vision like object recognition [21]. The method of Oriented FAST [22] and Rotated BRIEF [23], where Fast algorithm is used for detecting keypoint, and Brief descriptor, with modifications to enhance performance applying Harris Corner [24], operates by comparing the same set of smoothed pixel pairs for each local patch that it describes. Parameters like area, circularity, convexity, and inertia was customized to evaluate patterns in the femoral head images during the segmentation procedure.

*MSER Detector:* Maximally Stable Extremal Region (MSER) extrator proceeds by sorting pixels by intensity, then each pixel are marked and the list of growing and merging connected components and their areas is maintained using the union-find algorithm [25]. The method was used to detect coordinates of bone structures.

*KAZE Detector:* The multiscale methods creates the scale space of an image by filtering the original image with an appropriate function over increasing time or scale [26]. The space is created by efficient additive operator splitting and variable conductance diffusion. The method was used to abstract the image by automatically detecting features of interest at different scale levels. For each of the detected features an invariant local description of the image could be obtained.

*BRISK Detector:* Binary Robust Invariant Scalable Keypoint (BRISK) detector relies on an easily configurable circular sampling pattern from which it computes brightness comparisons to form a binary descriptor string [27].

All methods implemented within the segmentation software has its custom parameters and could be changed by the user shows in figure 1c. The toolkit of image processing was used to set the best approach for the autosegmentation of femural head.

*4. Auxiliary Structure*

An auxiliary algorithm was developed for loading, saving, and contouring the images of pelvic CT scan. It was composed by browse file, show image, remove couch, save image, and overlap contour. The function of each structure is showed below. The figure 2 shows the flowchart of the femural head automatic delineation.

*Browse File:* Asks the user to choose a single or set of images and stores for analysis.

*Show Image:* Shows the original image and the last modified one for user's analysis.

*Remove Couch:* It is composed by a hard-coded algorithm to remove couch (table) from images of CT scan, makes bone isolation, and appropriated autosegmentation.

*Save Image:* Asks where the user wants to save the last modified image.

*Overlap Contour:* Merges the segmented image with the original image.

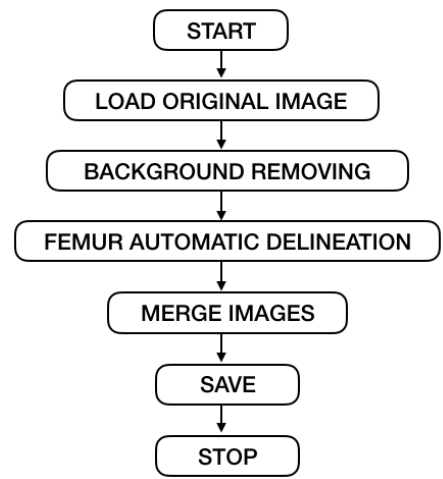

Fig. 2. Flowchart of the femural head automatic delineation.

*B. Software Testing*

The algorithm developed was applied to delineate the femural head of pelvic CT scan. Eleven 3D Radiotherapy Planning (Eclipse®, Varian, USA) of prostate patients clinically treated from the database of the Sao Paulo Hospital (5 patients) and AC Camargo Cancer Center (6 patients) were used in this study. All experimental procedures were carried out in accordance with the World Medical Association International Code of Medical Ethics and were approved by the local Ethics Committee.

*C. Data Analysis*

The automatic delineation was compared with manual and approved delineation through blind test evaluated by seniors radiation oncologists from both hospitals (6 from AC Camargo Cancer Center and 3 from Sao Paulo Hospital). The survey showed the femural delineation to doctors without informing which ones were done automatically or manually.

The survey one was composed by 60 random images of the manual and automatic delineation placed side by side, and the doctors (n=9) were asked to choose which segmented image would be approved (both, 1st delineation, 2nd delineation, or none). The survey two was composed by 10 complete femural delineation (36 images each), where 5 delineation were done manually and approved by radiation oncologists and another 5 ones were done automatically by software. The doctors (n=6) were asked to evaluate if the delineation needs some changes (None, small change, medium change, or large change). The qualitative and quantitative research was employed for the purpose to analise the software performance and the intra- and inter-observer variations due the factors such as user preferences and experience of the radiation oncologists.

The delineation data were divided in three femural region (proximal, medial, and distal) aiming for a better analysis of the software performance. Proximal region consisted of the lesser trochanter, Medial region consisted between the end of lesser trochanter and the end of greater trochanter (the region of the femur neck), and Distal region consisted by the femural head.

## III. RESULTS AND DISCUSSION

The software developed presents the ability to delineate the entire femural head for 3D Radiotherapy Planning. It identified and started the automatic delineation at beginning of less trochanter and stoped in the end of femural head properly. The sample of automatic delineation results for proximal, medial, and distal femur region can be seen in Fig. 3. In a visual inspection and qualitative analyses, the software shows a great performance for the proximal femur region where the high contrast between bone and soft tissue occur. For medial and distal femur region the software presents less precision than the proximal part. The less autosegmentation accuracy occur where there are low contrast between bones structures like the anatomical union between the femural head and acetabulum at the pelvic region. This is a challenge for any autosegmentation algorithm and it could be satisfactory achieved by the present software.

Fig. 4 shows the evaluation of seniors radiation oncologists on the blind comparison between the manual and automatic femur delineation on the survey one and survey two. The results were evaluated separately for proximal, medial, and distal femur region.

Fig. 4a. shows that the automatic delineation had the same performance than the manual delineation in the 89% of the CT slices and in 9.5% the automatic delineation was better than the manual for the proximal femur region. This means a medical delineation approval for 98.5% of automatic delineation and a refusal of 10.0% on the manual delineation done by the same seniors radiation oncologists.

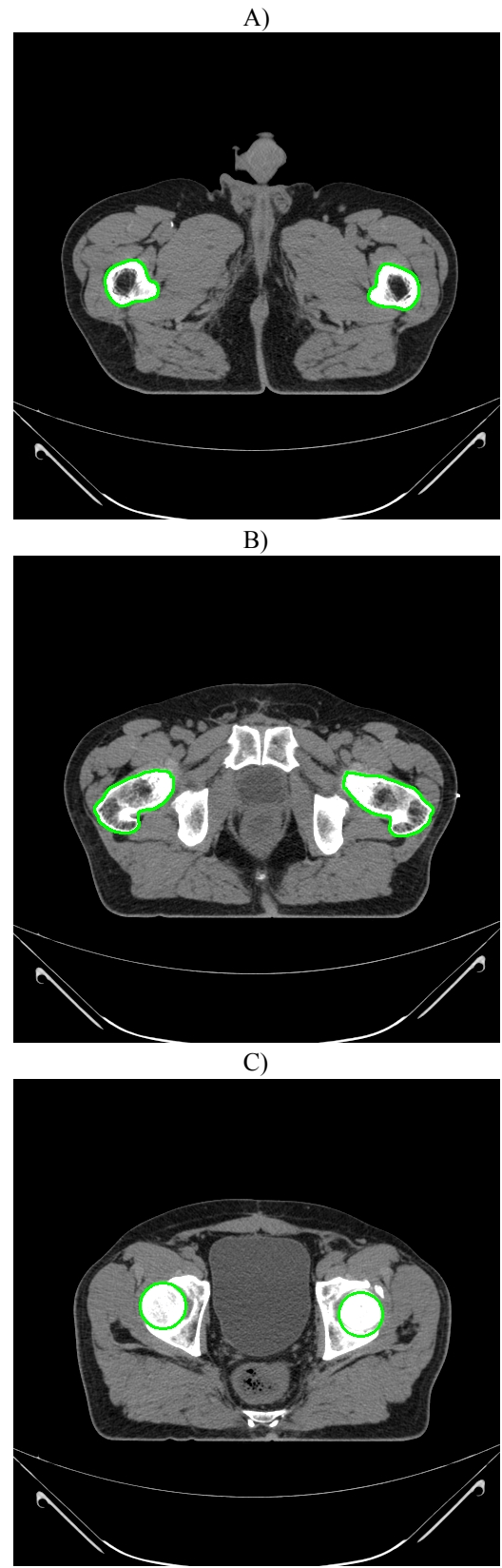

Fig. 3. Autosegmentation of femural head at proximal (A), medial (B), and distal (C) region. The green area represent the automatic delineation. The delineations showed are a random sample of all autosegmentation obtained for all patients (n=11).

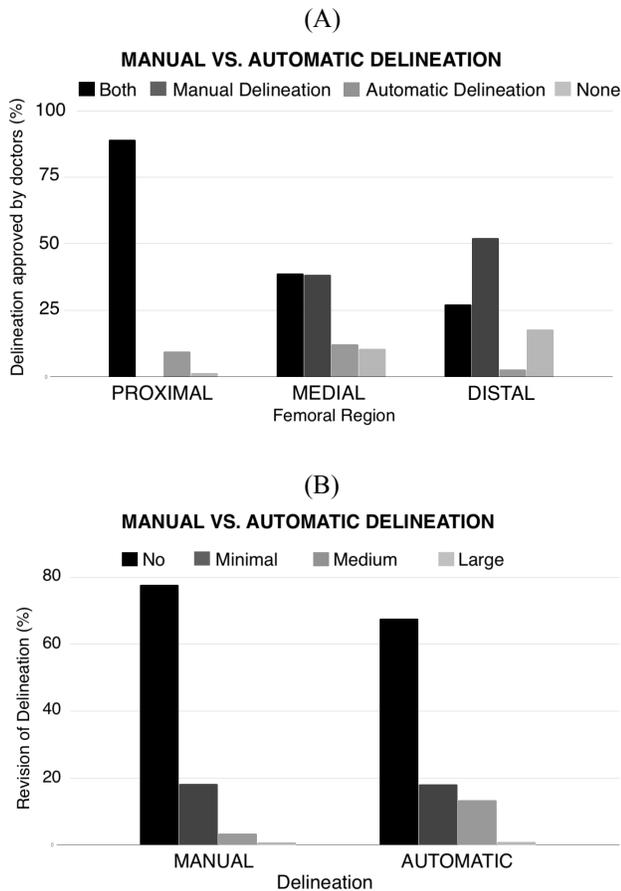

Fig. 4. Comparison between manual (approved delineation) and automatic femural head delineation evaluated through blind test by a panel of seniors radiation oncologist from two separately cancer center hospitals. A) Survey composed by 60 random images of the manual and automatic delineation placed side by side for doctors (n=9) approval. B) Survey composed by 10 complete femural delineation (36 images each), where 5 delineation were done manually and another 5 ones were done automatically by software. The doctors (n=6) were asked to evaluate if the delineation needs some changes.

Fig. 4a. shows that the automatic delineation was approved in 51.1% and 30.0% of the medial and distal femur region respectively. It is important to highlight the fact that the manual approved delineation was refused by the radiation oncologists in 22.8% and 17.8% of the medial and distal femur region respectively. According Pekar *et al.* [3], manual segmentation is considered the "gold standard", but cannot be considered the perfect ground truth, due to inter-observer variability.

Fig. 4b. shows that among all femural delineation (manual and automatic) analyzed separately by the survey two, 77.8% of manual delineation and 67.6% of automatic delineation do not need of any contouring correction, approximately 18% of manual and automatic delineation need of a minimal correction, 0.3% of manual delineation and 13.4% of automatic delineation need of a medium correction, and less than 1.0% of manual and automatic delineation need of large correction. The results shows that the software developed has a similar perfomance compared with delineation by a radiation oncologist. Whereas that it is preliminary software version,
improvements can still be made in terms of precision and accuracy, specially related on delineation of the medial and distal femur region, where the software performance can be improved.

## IV. CONCLUSION

The main issue regarding femural head delineation was addressed in this paper. A totally visual Application Program Interface was proposed, based on several types of pre- and post-processing image segmentation, and test results showed excellent performance. The automatic delineation was reproducible in all patient and its performance was similar to manually delineation made and evaluated by a panel of senior radiation oncologist. As a preliminary software version, it needs of precision and accuracy improvements on the medial and distal femural region. While, the proximal region showed that the automated delineation may have more accuracy than a clinical expert's eye. Although, the goal of the automated procedures is to relieve the physician of time-consuming tasks, as repetitive organs at risk delineation, and improve their personal planning to gain more time just for a delineation review on organs at risks and focus in making a target delineation. The software developed shows the great potential to be employed for automatic segmentation of other anatomical body region. The results that have been achieved should be seen in a very positive light for further reduce the processing time and enable clinical routine use.

ACKNOWLEDGMENT

The authors would like to thank Dr. Rodrigo Souza Dias and Dr. Antonio Cassio de Assis Pellizzon from, respectively, Sao Paulo Hospital and AC Camargo Cancer Center, and a special thanks to Dr. Luiz Juliano Neto. This work is part of the agreement between Varian Medical Systems and Brazilian Ministry of Health for the expansion of radiotherapy in Brazil.